%% file: main.tex
\documentclass{article}

\usepackage[preprint]{corl_2026} 
 \usepackage{amsmath}
 \usepackage{enumitem}
 \usepackage{graphicx}
 \usepackage{natbib}
 \usepackage{graphicx}
 \usepackage{amsmath,amssymb} 
 \usepackage{wrapfig}
 \usepackage{multirow}
 \usepackage{booktabs}
 \usepackage[utf8]{inputenc}
 \usepackage{subcaption}
 \usepackage{amsthm}
 \usepackage[dvipsnames]{xcolor}
\usepackage{subcaption}
\usepackage{hyperref}
\usepackage[table]{xcolor}
\usepackage{booktabs}
\usepackage[most]{tcolorbox}
\usepackage{authblk}
\usepackage{xcolor}
\usepackage{tikz}
\usetikzlibrary{calc}

\title{\textcolor{RedOrange}{EAGOR}: \textcolor{RedOrange}{E}mbodied re\textcolor{RedOrange}{A}sonin\textcolor{RedOrange}{G} in \textcolor{RedOrange}{O}mni-di\textcolor{RedOrange}{R}ection\vspace{-10pt}}

\makeatletter
\def\AB@affilnote{}
\makeatother

\author{
Shriram Damodaran$^{1,*}$,
Soumyaratna Debnath$^{1,*}$,
Yan Wu$^{2}$,
Wei-Yun Yau$^{2}$,
Lin Wang$^{1,\dagger}$
\\
$^{1}$EmPACT Lab, NTU Singapore
\qquad
$^{2}$Institute for Infocomm Research, A*STAR Singapore
\\
$^{*}$Equal contribution.
\qquad
$^{\dagger}$Corresponding author: \texttt{linwang@ntu.edu.sg}
}

\definecolor{darkgreen}{rgb}{0.05, 0.60, 0.18}
\definecolor{cvprBlue}{rgb}{0.21,0.49,0.74}
\definecolor{DeepPink}{rgb}{1,0.078,0.576}
\definecolor{SuppColor}{rgb}{0.9961,0.4353,0.3686}
\definecolor{GRAY}{rgb}{0.4, 0.4, 0.4}
\hypersetup{
    breaklinks=true,
    colorlinks=true,
    allcolors=DeepPink
}

\begin{document}

\begin{tikzpicture}[remember picture, overlay]

\node[anchor=north west, inner sep=0pt]
at ($(current page.north west)+(0.72in,-0.43in)$)
{
    \includegraphics[
        width=3.0cm,
        trim=3 3 3 3,
        clip
    ]{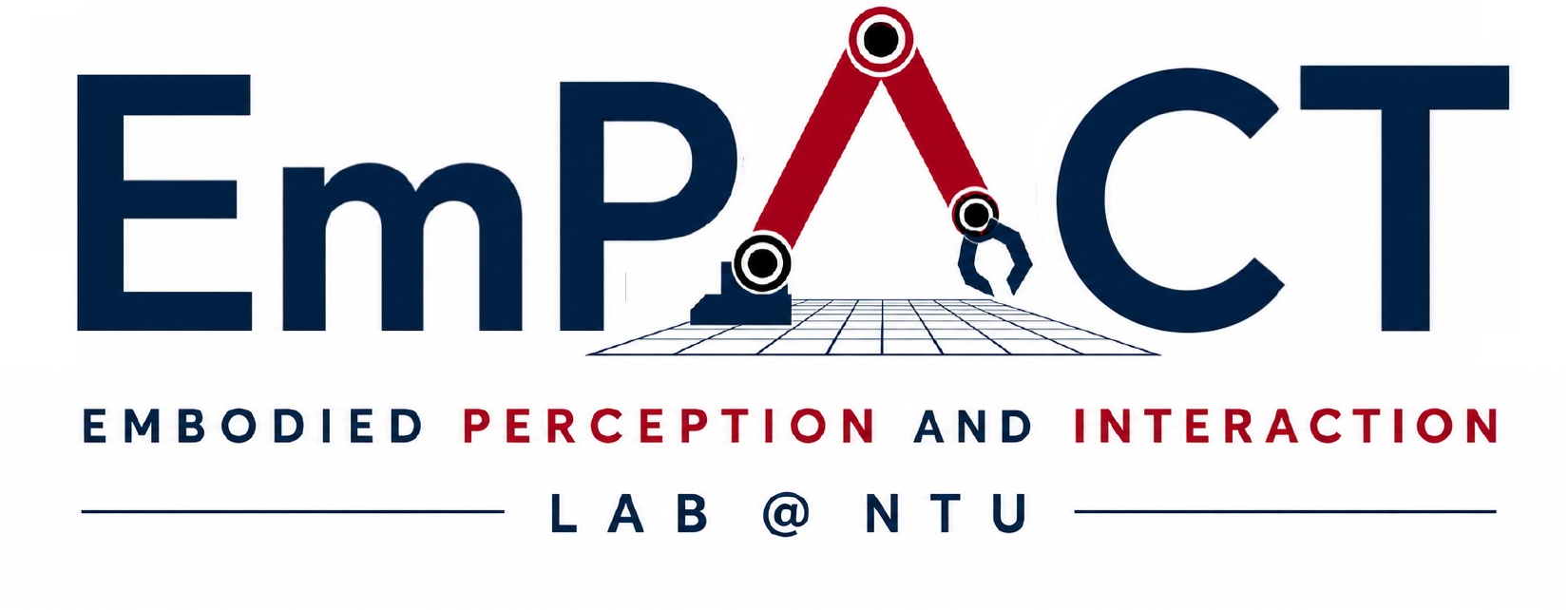}
};

\draw[
    black!20,
    line width=0.35pt
]
($(current page.north west)+(0.72in,-0.98in)$)
--
($(current page.north east)+(-0.72in,-0.98in)$);

\end{tikzpicture}

\maketitle

\begin{figure}[h!]
\label{sec:teaser}
    \centering
    \vspace{-27pt}
    \includegraphics[width=0.79\linewidth]{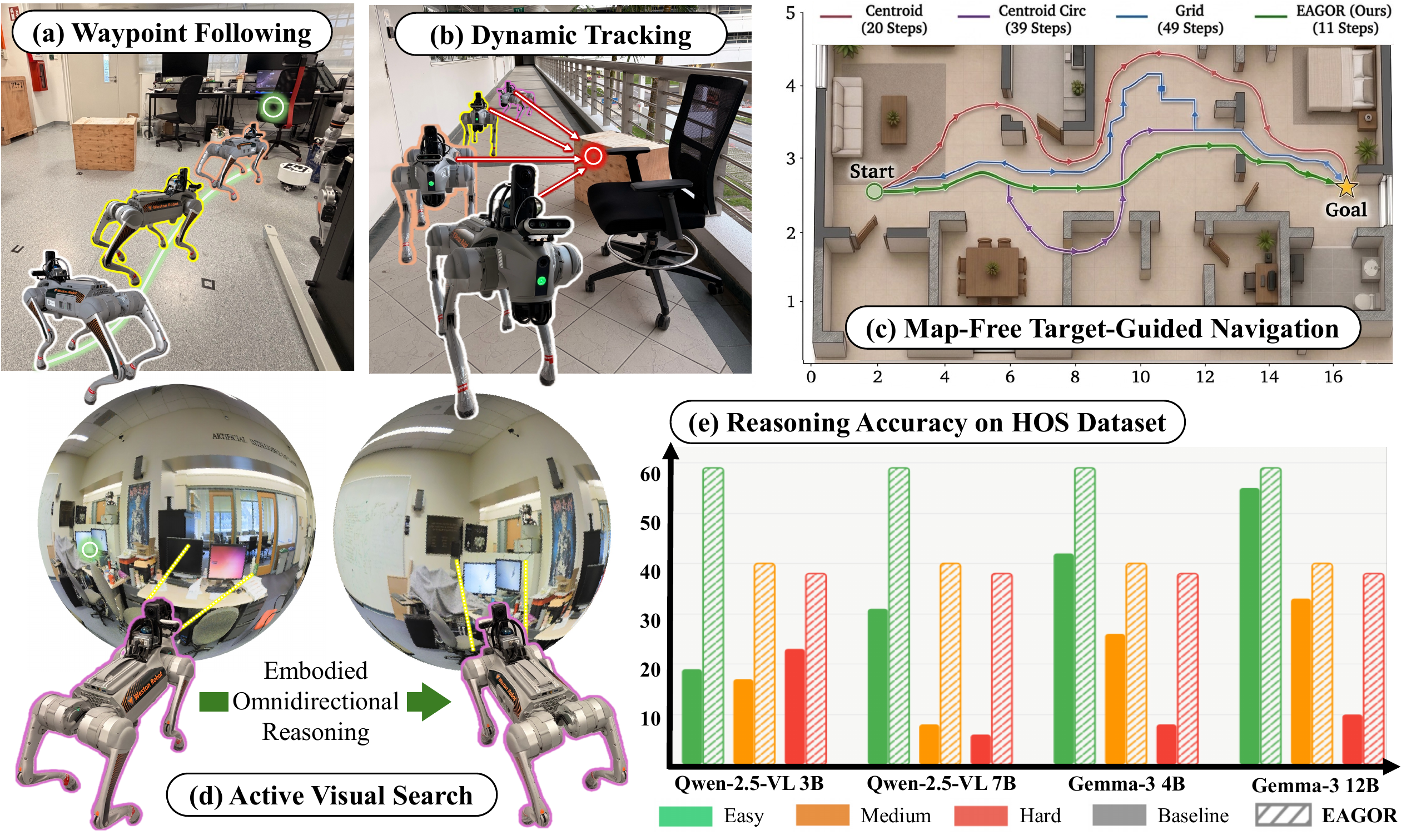}
    \vspace{-5pt}
    \caption{\footnotesize\textbf{\textcolor{RedOrange}{EAGOR}} for geometry-aware embodied omni-directional reasoning ($\bigcirc=$~Target)}
    \label{fig:teaser}
    \vspace{-12pt}
\end{figure}

\input{sections/abstract}
\input{sections/introduction}

\input{sections/related_works}
\input{sections/methodology_1}

\input{sections/experiments}

\input{sections/conclusion}

\clearpage
\bibliography{reference}  

\end{document}

%% file: sections/abstract.tex
\begin{abstract}
Omni-directional ($360^\circ$) cameras enable embodied agents with a wide and holistic view of their surroundings, making them advantageous for directional reasoning in embodied tasks, such as navigation, and object search. 
Existing Vision Language Models (VLMs) project $360^\circ$ data to 2D planar images via commonly used equirectangular projection (ERP) and process them with architectures designed for perspective images. However, they overlook the fact that $360^\circ$ data is fundamentally spherical, with each pixel encoding a direction relative to the agent's view.
As a result, existing methods often produce direction estimates that are inconsistent under camera view transformation due to agent motion. 
This becomes critical in map-free navigation, where the agent must continuously update it's direction to a target in its egocentric frame as it navigates in $360^\circ$ environment.
In this paper, we propose \textbf{EAGOR}, a \textbf{training-free} and \textbf{geometry-aware} framework for embodied $360^\circ$ directional reasoning. \textbf{Our key idea} is to reframe the dynamic agent-to-target directional relationship from $360^\circ$ observations as \textit{recursive Bayesian estimation on the sphere}, rather than as pixel-coordinate prediction in an ERP image. 
EAGOR therefore maintains a continuous, geometrically correct belief to a target direction on the sphere, and propagating it equivariantly under agent motion, without training the backbone VLMs. 
To realize this formulation, we introduce a \textbf{Spherical Harmonic Belief Field (SH-BF)}, whose spherical harmonic representation provides a globally defined and rotation-aware basis for target direction estimation directly on the spherical manifold.
As a result, EAGOR avoids the seam discontinuities, latitude distortions, and interpolation errors caused by the ERP.
We evaluate EAGOR on two benchmark datasets and real-world experiments with a legged robot for multiple directional reasoning tasks.
EAGOR consistently outperforms the baselines, achieving average relative gains of $\mathbf{+34.4\%}$ and $\mathbf{+45.6\%}$ on HOS and OSR-Bench, respectively, for active visual search.
Notably, EAGOR improves navigation success by $\mathbf{+14.6\%}$, reducing steps counts by $\mathbf{17.7\%}$ and mean angular error by $\mathbf{24.5\%}$.
\end{abstract}
\vspace{-10pt}
\keywords{Embodied $360^\circ$ Reasoning, Map-Free Navigation, Visual Search}

%% file: sections/introduction.tex
\section{Introduction}
\vspace{-15pt}
Directional reasoning -- the fundamental ability to estimate, track, and update spatial vectors relative to the agent's own physical posture -- is a cornerstone of embodied AI. Just as humans rely on a continuous, multi-sensory awareness of their surroundings to track objectives even when looking away, a physical agent, like robot, must accurately compute the egocentric direction of a target and dynamically maintain it during ego-motion \cite{das2017embodiedquestionanswering,krantz2020navgraphvisionandlanguagenavigationcontinuous, manh2025mind}.
\vspace{-3pt}

Omni-directional (\(360^\circ\)) cameras capture full surrounding environment with a $180^\circ \times 360^\circ$ field-of-view (FoV), enabling agents to observe targets, landmarks, and obstacles in all directions
\cite{853799,anderson2018visionandlanguagenavigationinterpretingvisuallygrounded}.
This makes them advantageous for directional reasoning in embodied AI tasks, such as map-free navigation and active visual search. 
In practice, spherical data is transmitted into 2D planar representations via equirectangular projection (ERP), \textit{a.k.a.} panorama to preserve
omni-directional information.  
However, ERP introduces seam discontinuities and latitude-dependent distortions, which violate the Euclidean assumptions built into the recent popular Vision-Language Models (VLMs), \textit{e.g.}, \cite{bai2023qwentechnicalreport}. Consequently, it causes spatial relations and direction estimates to be inconsistent due to camera transformations or agent motion~\cite{dong2023panocontextformerpanoramictotalscene,10.1007/978-3-030-01270-0_43,10.1007/978-3-030-01240-3_32,su2018learningsphericalconvolutionfast,chou2020visualquestionanswering360deg, debnath2026adaptopass}. 
\vspace{-3pt}

Crucially, omni-directional reasoning represents the surrounding scene in terms of \textbf{viewing directions around the agent}, rather than positions on a flat image plane. 
For embodied omni-directional reasoning, the relevant output is therefore not only the \textbf{location} of a target, but also the \textbf{egocentric direction} the agent should turn, move, or search toward \cite{chaplot2020objectgoalnavigationusing}.
As illustrated in Fig.~\ref{fig:intro-fig}, projection-based representations can accumulate interpolation errors under camera rotation, causing the predicted target direction to drift over time.
This reveals a representation gap in omni-directional reasoning. While VLM's provide strong semantic observations in image space, 
the underlying geometry must be handled in a representation that respects the observer-centered geometry of the sphere. 
\vspace{-3pt}

For action, the agent needs a continuous egocentric direction to the target that remains consistent under motion, not merely a 2D pixel location in the panorama.
A suitable framework should therefore satisfy two requirements:
\textbf{(a)} represent target direction probabilistically on the sphere (Fig.~\ref{fig:teaser}\textcolor{DeepPink}{d}), and
\textbf{(b)} propagate this state consistently as the agent moves and rotates (Fig.~\ref{fig:teaser}\textcolor{DeepPink}{a,b}).
This motivates our research question:
\textit{Can VLMs perform geometry-aware omni-directional reasoning through a continuous belief representation over panoramic observations without task-specific training?}
\vspace{-3pt}

To answer this, we introduce \textbf{EAGOR} 
(\textbf{E}mbodied re\textbf{A}sonin\textbf{G} in \textbf{O}mni-di\textbf{R}ection), 
a \textbf{training-free} and \textbf{geometry-aware} embodied omni-directional reasoning framework. 
EAGOR decouples semantic perception from geometric state estimation.
Instead of directly asking detectors or vision-language models to predict the final target direction, EAGOR uses them to extract target-conditioned evidence from the current panorama. The omni-directional inference is remodeled as belief estimation on the sphere.
This evidence is lifted from projected image coordinates to viewing directions on the sphere and treated as an observation likelihood over possible target directions.
We model this likelihood with a \textbf{Spherical Harmonic Belief Field (SH-BF)}, representing continuous directional belief in the SH domain (Sec.~\ref{sec:sh-bf}). 
This enables precise likelihood estimation required for accurate omni-directional reasoning
while avoiding the artifacts caused by ERP, as shown in Fig.~\ref{fig:teaser}\textcolor{DeepPink}{d}. 
\vspace{-3pt}

We evaluate EAGOR across multiple embodied omni-directional reasoning tasks: \textit{Waypoint Following} and \textit{Map-free Navigation} in Habitat-Sim; \textit{Active Visual Search} on HOS and OSR-Bench; and real-world \textit{Dynamic Tracking} on a legged robot.
Notably, for \textit{Active Visual Search} (Fig.~\ref{fig:teaser}\textcolor{DeepPink}{e}), EAGOR achieves up to $\sim1.5\times$ gain on HOS and over $\mathbf{35\%}$ relative improvement with a $\mathbf{10\times}$ smaller backbone on OSR-Bench. 
For \textit{Map-free Navigation} (Fig.~\ref{fig:teaser}\textcolor{DeepPink}{c}), it reduces temporal inconsistency by $6.5\times$. 
Real-world results further demonstrate generalization of our training-free approach.

\vspace{-3pt}
In summary, our contributions are:
\textbf{(I)} We identify a critical representation gap in embodied omni-directional reasoning -- 
robotic action requires a motion-consistent directional belief on the sphere (Fig.~\ref{fig:intro-fig}).
\textbf{(II)} We propose EAGOR, a training-free and geometry-aware framework that models VLM-based semantic evidence as likelihood on the sphere for accurate omni-directional reasoning (Fig.~\ref{fig:methodology}). 
\textbf{(III)} We introduce SH-BF for representing continuous and rotation-consistent directional belief propagation under robot motion (Sec.~\ref{sec:sh-bf}).
\textbf{(IV)} We validate EAGOR across multiple embodied omni-directional reasoning tasks and benchmarks, together with real-world validation on a legged robot, showing superior reasoning performance for navigation and visual search (Sec.~\ref{sec:result}).

%% file: sections/related_works.tex
\vspace{-10pt}
\section{Related Works}
\vspace{-10pt}

\begin{figure*}[t!]
    \centering
    \vspace{-10pt}
    \includegraphics[width=0.95\linewidth]{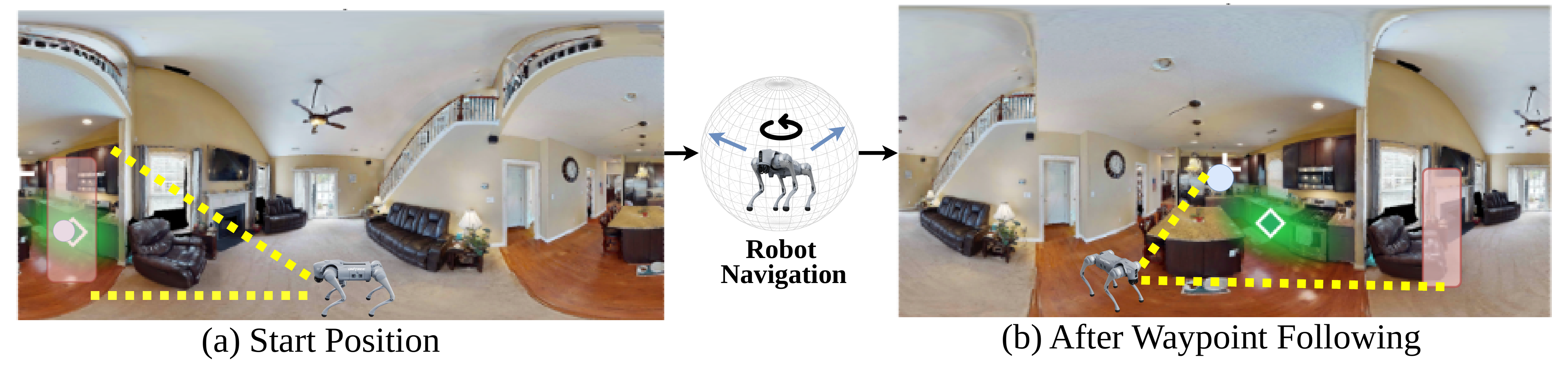}
    \vspace{-10pt}
\caption{\textbf{Qualitative results for \textit{Waypoint Following} task.} 
\textbf{(a)} Initial target-direction estimate. 
\textbf{(b)} Directional estimates after viewpoint changes during waypoint following. 
\textit{Qwen2.5-VL + EAGOR} ({\color{ForestGreen}{\Large$\diamond$}}) consistently preserves the target direction under agent motion, whereas the centroid ({\color{black}$\bigcirc$}) and grid-based ({\color{red}$\square$}) baselines drift due to accumulated ERP interpolation errors.}
    \label{fig:intro-fig}
    \vspace{-12pt}
\end{figure*}

\textbf{Omni-directional Perception for Embodied Navigation.}
Early approaches address ERP distortions by decomposing panoramas into perspective 
crops and applying standard pretrained 
models~\cite{yun2021panoavqagroundedaudiovisualquestion,chou2020visualquestionanswering360deg,
eder2020tangentimagesmitigatingspherical}, while recent methods incorporate 
panorama-aware attention, spherical positional encodings, or large-scale panoramic 
supervision~\cite{benny2024sphereuformerushapedtransformerspherical,
zhang2022bendingrealitydistortionawaretransformers,unlu2023sphericalpositionencodingtransformers,
li2025da2depthdirection,wang2026panoworldspatialsupersensing360circ}; yet benchmarks such as OSR-Bench 
and ODI-Bench reveal that MLLMs still struggle with
reference-frame transformation~\cite{yang2026odibenchmllmsunderstandimmersive,
dongfang2025multimodallargelanguagemodels,liu2025spatialreasoningmultimodallarge,
zheng2025multimodalspatialreasoninglarge}, indicating that semantic recognition does 
not guarantee geometric correctness: \textit{a model may identify the target while 
estimating its direction incorrectly.}
These limitations become critical in embodied applications such as navigation 
and active visual search, where the agent must dynamically maintain its egocentric 
target direction under motion.
Instead of reasoning directly over sphere, prior methods typically depend on global reference frames, prebuilt maps, per-frame decisions from panoramic crops, or waypoint prediction in planar coordinates~\cite{zhao2024imaginenavpromptingvisionlanguagemodels,jin2025panonavmaplesszeroshotobject,yokoyama2023vlfmvisionlanguagefrontiermaps,wang2023gridmmgridmemorymap,wang2023dreamwalkermentalplanningcontinuous,zhang2025uninavidvideobasedvisionlanguageactionmodel}. Consequently, their directional estimates remain tied to representations that distort under egocentric rotation.
\textit{Our work therefore treats directional reasoning not as a recognition problem to 
be solved per frame, but as a geometric state estimation problem to be maintained 
continuously on the sphere.}
\vspace{-2pt}

\textbf{Probabilistic Evidence Accumulation for Direction Estimation.}
A deeper limitation of existing reasoning frameworks is they treat VLM spatial outputs as deterministic coordinate predictions, discarding the underlying uncertainty and using a single noisy or incorrect observation as a definitive answer~\cite{yu2025thinking360deghumanoidvisual,jin2025panonavmaplesszeroshotobject,wang2026panoworldspatialsupersensing360circ}. Rather than explicitly modeling this uncertainty, prior approaches typically rely on repeated querying or confidence-based filtering~\cite{kendall2016modellinguncertaintydeeplearning,baumann2026posthocprobabilisticvisionlanguagemodels}.
In embodied settings, direction estimation is inherently sequential: as the agent moves, the agent-to-target direction must be updated under egocentric rotation. 
 We instead treat observations as probabilistic directional likelihood and accumulate them recursively over time \cite{Kurz_2016,Peretroukhin_2020}, forming a continuous belief over possible target directions on the sphere. This allows multiple plausible directions to be maintained and refined over time as new evidence is accumulated. Unlike prior probabilistic approaches~\cite{shao2025pointcapturinguncertaintyadaptive,zangeneh2023probabilisticframeworkvisuallocalization} operating in 2D planar image space, our framework maintains and updates this belief directly on the sphere .
\vspace{-2pt}

\textbf{Spherical Harmonic Representations.}
Spherical harmonics (SH) provides a equivariant representation for signals on $\mathbb{S}^2$ 
and have been widely adopted for equivariant feature learning, 
omni-directional learning tasks~\cite{esteves2018learningso3equivariantrepresentations,
10.1007/978-3-030-01240-3_32,liao2023equiformerequivariantgraphattention,Lee_2025_CVPR}, such as localization~\cite{10.1007/978-3-030-69538-5_21}, 
depth estimation~\cite{Lee_2025_CVPR}, and scene 
understanding~\cite{10.1007/978-3-030-01240-3_32}.
However, all these works take SH representations as learned task-specific 
features~\cite{esteves2018learningso3equivariantrepresentations,liao2023equiformerequivariantgraphattention}, 
or encoded analytic signals~\cite{Lee_2025_CVPR}.
In this work, we leverage SH as a belief representation for recursive Bayesian estimation which incorporates noisy, open-vocabulary observations as directional likelihoods. 
\textit{Our proposed SH-BF realizes the  belief representation by connecting frozen VLM evidence to recursive 
spherical belief estimation, enabling geometrically consistent agent to target directional representation across embodied navigation and active visual search.}



%% file: sections/methodology_1.tex
\vspace{-10pt}
\section{Methodology}
\vspace{-10pt}
\label{sec:methodology}

\begin{figure*}[t!]
    \centering
    \includegraphics[width=0.95\textwidth]{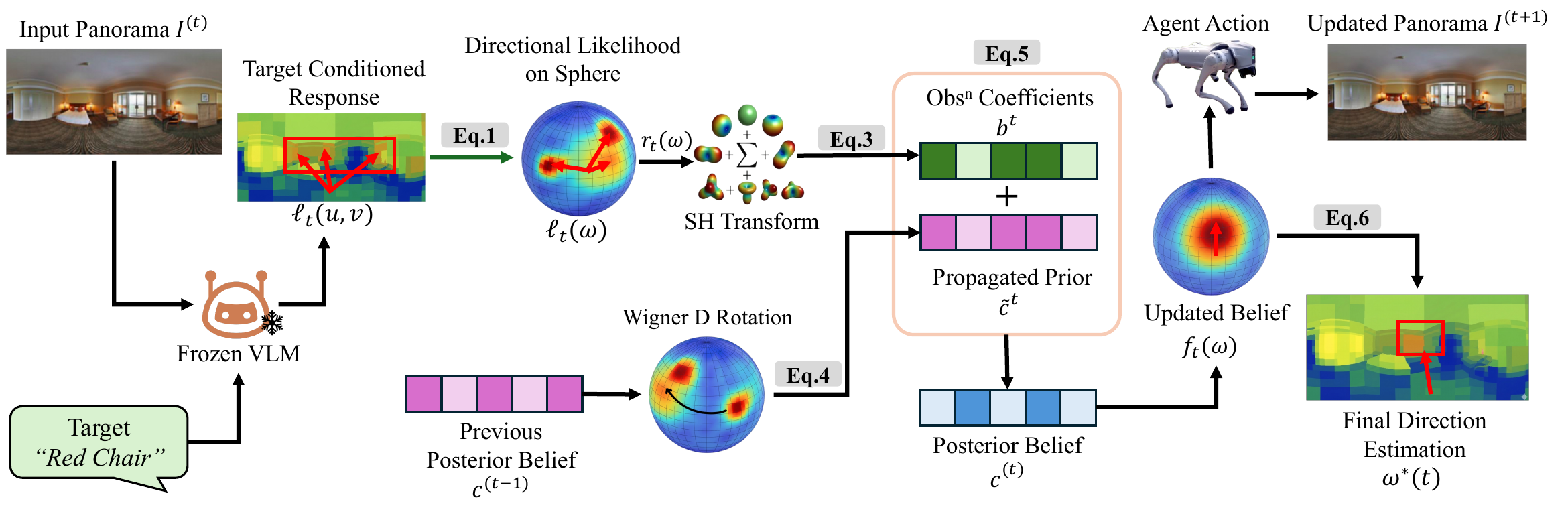}
    \vspace{-10pt}
    \caption{\textbf{Overview of EAGOR framework.} \textit{See main texts of Sec.3 for details.}}
    \label{fig:methodology}
    \vspace{-18pt}
\end{figure*}

\textbf{Overview.} Fig.~\ref{fig:methodology} illustrates the overview of the EAGOR framework.  
The key idea is to \emph{decouple semantic perception from geometric state estimation}. Rather than predicting target direction as a pixel coordinate in an ERP image, EAGOR treats the agent-to-target direction as a continuous belief on the sphere. 
Given a panoramic observation $I^t$ and a target query, a frozen VLM provides semantic evidence, which is lifted to the spherical domain and fused into a directional belief. To realize this, we introduce the \textbf{Spherical Harmonic Belief Field (SH-BF)}, which represents belief on $\mathbb{S}^2$ and supports consistent evidence accumulation, equivariant propagation under agent rotation, and MAP direction estimation. In short, the VLM identifies \emph{what} to look for, while SH-BF maintains \emph{where} the target lies relative to the agent at time $t$.

\vspace{-10pt}
\subsection{From panoramic observations to directional likelihoods}
\vspace{-8pt}

At each timestep \(t\), the agent receives a full \(360^\circ\) panoramic
image \(I_t\) and a target query specifying the object of interest.
The task is to estimate the egocentric agent-to-target direction
\(\boldsymbol{\omega}^{*}_{t}\in\mathbb{S}^2\), represented as a unit
vector in the viewer-centric frame.
\vspace{-3pt}

Given \(I_t\) and the query, the VLM produces a target-conditioned response
map \(\ell_t(u,v)\in[0,1]\) over the ERP image, where each score indicates
how likely the target is to appear at image location \((u,v)\).
As shown in Fig.~\ref{fig:methodology}, the response map may contain
multiple candidate regions in image space, corresponding to several
plausible target directions in the current panorama.
We interpret \(\ell_t\) as a directional observation likelihood field under
an explicit modelling assumption: higher VLM attention at a direction
correlates with a higher probability of the target being present there.
This assumption is treated as a stated model rather than an implicit
identification, and is validated empirically via a calibration analysis on
held-out episodes (\textcolor{Bittersweet}{\textit{Sec.~1.1 of Suppl.~Mat.}}).
\vspace{-3pt}

For an image of width \(W\) and height \(H\), we map pixel coordinates to
azimuth and elevation by \(\theta=2\pi u/W-\pi\) and \(\phi=\pi/2-\pi v/H\),
giving the unit direction
{\setlength\abovedisplayskip{3pt}
\setlength\belowdisplayskip{3pt}
\begin{equation}
\boldsymbol{\omega}(u,v)
=
(\cos\phi\cos\theta,\,\cos\phi\sin\theta,\,\sin\phi).
\end{equation}}
Using this mapping, \(\ell_t(u,v)\) is lifted to a directional likelihood
field \(\ell_t(\boldsymbol{\omega}):\mathbb{S}^2\rightarrow[0,1]\).
As shown in Fig.~\ref{fig:methodology}, lifting the response to the sphere
converts image-space candidates into directional likelihood peaks on
\(\mathbb{S}^2\), with their relative intensities reflecting the VLM's
confidence.
The response map is normalized to \([0,1]\) before the log transform.
To combine evidence across multiple timesteps additively, we convert the
directional likelihood into a log-likelihood field
\(r_t(\boldsymbol{\omega})=\log(\ell_t(\boldsymbol{\omega})+\varepsilon)\),
where \(\varepsilon\) avoids numerical instability.
We treat \(r_t(\boldsymbol{\omega})\) as the current observation in the
Bayesian estimator — the directional evidence provided by the current
panorama at time \(t\). Because a single-frame observation can be noisy
due to false detections, occlusion, or ERP distortion, EAGOR does not
estimate the final target direction from one image alone.
Instead, it treats target localization as a recursive state estimation
problem on \(\mathbb{S}^2\), where each panorama provides a new directional
observation and the final estimate is obtained by accumulating evidence
over time.

\vspace{-10pt}
\subsection{Spherical Harmonic Belief Field (SH-BF)}
\label{sec:sh-bf}
\vspace{-5pt}

Given the current directional observation \(r_t(\boldsymbol{\omega})\),
EAGOR maintains a persistent belief over possible target directions on the
sphere. We denote this belief as a continuous field
\(f_t:\mathbb{S}^2\rightarrow\mathbb{R}\), representing the accumulated
log-posterior score that the target lies in direction
\(\boldsymbol{\omega}\):
\(f_t(\boldsymbol{\omega})\approx\log p(\boldsymbol{\omega}^{*}
=\boldsymbol{\omega}\mid z_{1:t},a_{1:t})\),
where \(r_t(\boldsymbol{\omega})\) is the directional evidence extracted
only from the current panorama at timestep \(t\), and \(f_t(\boldsymbol{\omega})\)
is the running belief obtained by recursively combining information over
time. Refer to \textcolor{Bittersweet}{\textit{Sec.~1.2 of Suppl.~Mat.}}
for details.
\vspace{-3pt}

To support dynamic belief updates under agent motion, the representation
must satisfy two requirements: (1)~it must accumulate evidence additively
in log space, and (2)~it must transform equivariantly under egocentric
rotation. We represent the belief in the real spherical harmonic (SH) basis,
which provides a continuous global representation on \(\mathbb{S}^2\) and
allows both operations to be performed directly in coefficient space without
ERP distortions:
{\setlength\abovedisplayskip{1pt}
\setlength\belowdisplayskip{1pt}
\begin{equation}
f_t(\boldsymbol{\omega})
=
\sum_{\ell=0}^{L}
\sum_{m=-\ell}^{\ell}
c_{\ell m}^{(t)}
Y_{\ell}^{m}(\boldsymbol{\omega}),
\label{eq:sh_belief}
\end{equation}}
where \(\{c_{\ell m}^{(t)}\}\) are the belief coefficients and \(L=7\) is
the SH bandlimit, spanning \(64\) coefficients and resolving angular
features at \({\approx}25^\circ\) — within the single-frame observation
noise floor and the \(\pm30^\circ\) task success criterion — while avoiding
the severe sidelobe ringing that arises at higher bandlimits
(\textcolor{Bittersweet}{\textit{Suppl.~Mat.~Sec.~B.1}}).
The observation \(r_t(\boldsymbol{\omega})\) is projected onto the same
SH basis \(Y_{\ell}^{m}\) to obtain observation coefficients
\(\mathbf{b}^{(t)}\), encoding evidence from the current panorama.
All subsequent operations — rotation propagation, translation advection,
belief update, and direction decoding — are carried out entirely in
coefficient space.
\vspace{-3pt}

\begin{figure*}[t!]
    \centering
    \vspace{-10pt}
    \includegraphics[width=0.88\linewidth]{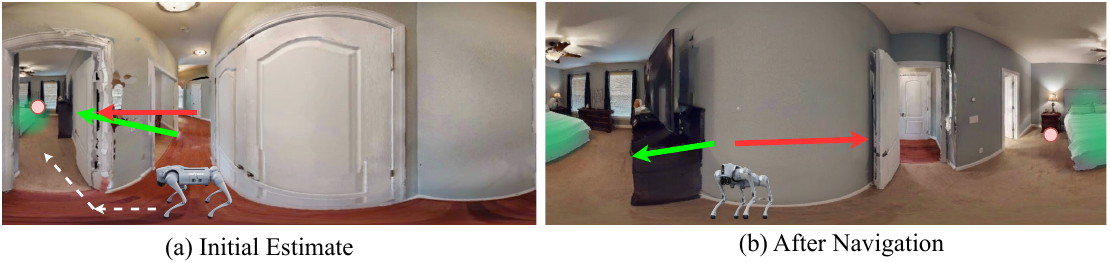}
    \vspace{-5pt}
    \caption{\textbf{Qualitative results for EAGOR using Qwen2.5-VL.}
    Direction reasoning during seam crossing (target: bed) where
    ${\color{ForestGreen}\rightarrow}$ indicates the EAGOR predicted direction,
    while ${\color{red}\rightarrow}$ shows the erroneous direction produced by
    the centroid-based baseline. $\bigcirc$ marks the belief centroid.}
    \label{fig:qualitative}
    \vspace{-20pt}
\end{figure*}

As shown in Fig.~\ref{fig:methodology}, at the first timestep no prior
belief is available, so the first panorama directly initialises the
posterior: \(\mathbf{c}^{(1)}=\mathbf{b}^{(1)}\). For each subsequent
timestep \(t>1\), the posterior \(\mathbf{c}^{(t-1)}\) is carried forward,
propagated into the current viewer-centric frame, and combined with
\(\mathbf{b}^{(t)}\) to form the updated posterior \(\mathbf{c}^{(t)}\),
which becomes the prior input for timestep \(t+1\).
\vspace{-3pt}

\noindent\textbf{Current observation in SH space.}
As illustrated in Fig.~\ref{fig:methodology}, the belief update proceeds
through two branches. The upper branch converts the current panorama into
observation coefficients \(\mathbf{b}^{(t)}\). We project the log-likelihood
field \(r_t(\boldsymbol{\omega})\) onto the SH basis, yielding:
{\setlength\abovedisplayskip{1pt}
\setlength\belowdisplayskip{1pt}
\begin{equation}
b_{\ell m}^{(t)}
=
\int_{\mathbb{S}^2}
r_t(\boldsymbol{\omega})\,Y_{\ell}^{m}(\boldsymbol{\omega})\,d\Omega
=
\int r_t(\theta,\phi)\,Y_{\ell}^{m}(\theta,\phi)\cos\phi\,d\phi\,d\theta.
\end{equation}}
The \(\cos\phi\) term is the spherical area element, correcting the
latitude-dependent oversampling inherent in ERP images.
\vspace{-3pt}

\noindent\textbf{Rotation propagation.}
Before fusing the new observation, the previous posterior belief must be
aligned with the new orientation. Because the belief is maintained in the
viewer-centric frame, it must rotate together with the agent. The lower
branch of Fig.~\ref{fig:methodology} performs this: for an egocentric
rotation \(R_t\in SO(3)\), the previous posterior coefficients
\(\mathbf{c}^{(t-1)}\) are propagated into the current frame via the
Wigner-D matrices,
{\setlength\abovedisplayskip{1pt}
\setlength\belowdisplayskip{1pt}
\begin{equation}
\tilde{c}_{\ell m}^{(t)}
=
\sum_{m'=-\ell}^{\ell}
D^{\ell}_{mm'}(R_t)\,c_{\ell m'}^{(t-1)},
\label{eq:wigner}
\end{equation}}
which act \emph{exactly} on SH coefficients at every degree \(\ell\)
independently — introducing no interpolation or resampling error,
regardless of the semantics of the expanded function. For planar
navigation, \(R_t\) reduces to the yaw change \(\Delta\psi_t\).
\vspace{-3pt}

\noindent\textbf{Translation propagation.}
For translation, we exploit a
structural property of translational camera motion that is
\emph{independent of range}: for a translating observer, the
\emph{direction} in which every scene point's apparent direction shifts
on \(\mathbb{S}^2\) is fully determined by the translation axis
\(\hat{\mathbf{t}}\) alone, and only the \emph{magnitude} depends on
the unknown scene range. This is the classical spherical
Focus-of-Expansion (FOE) / Focus-of-Contraction (FOC) structure:
the flow vanishes at the two poles \(\pm\hat{\mathbf{t}}\) and is
maximal on the great circle orthogonal to \(\hat{\mathbf{t}}\).
We define the unit flow direction field
\(\hat{V}(\boldsymbol{\omega})=-\Pi_{\boldsymbol{\omega}}(\hat{\mathbf{t}})\),
the tangent-plane projection of \(\hat{\mathbf{t}}\) at each
\(\boldsymbol{\omega}\), and model translation as first-order advection
of the belief along \(\hat{V}\):
{\setlength\abovedisplayskip{1pt}
\setlength\belowdisplayskip{1pt}
\begin{equation}
\tilde{c}_{\ell m}^{(t)}\big|_{\mathrm{trans}}
=
\tilde{c}_{\ell m}^{(t)}
-
\eta_t
\left(\mathcal{A}_{\hat{\mathbf{t}}}\,\tilde{\mathbf{c}}^{(t)}\right)_{\ell m},
\label{eq:translation}
\end{equation}}
where \(\mathcal{A}_{\hat{\mathbf{t}}}\) is the Lie derivative along
\(\hat{V}\) — a precomputed, degree-banded linear operator on SH
coefficients (\textcolor{Bittersweet}{\textit{Suppl.~Mat.~Sec.~B.4}}) — and
\(\eta_t=\|\mathbf{t}_t\|/\rho_{\min}\) is a worst-case scalar bound
using only the agent body standoff \(\rho_{\min}\) (a fixed physical
constant, not a scene-content estimate). As \(\|\mathbf{t}_t\|\to 0\),
\(\eta_t\to 0\) and Eq.~\eqref{eq:translation} reduces exactly to the
identity, recovering pure rotation propagation as a special case.

Critically, the belief is maintained in the sensor frame throughout,
so \(\hat{\mathbf{t}}\) always points along the sensor-frame forward
direction, regardless of world heading — a direct consequence of
Wigner-D in Eq.~\eqref{eq:wigner} already absorbing all egocentric
orientation changes before Eq.~\eqref{eq:translation} is applied.
The translation update is applied selectively: when a fresh VLM
observation is available it re-anchors the belief at the current
geometry, rendering the translation correction unnecessary.


\noindent\textbf{Belief update.}
At the first timestep, the posterior is initialised from the first
observation directly. For subsequent timesteps, the posterior is obtained
by combining the propagated prior (Eqs.~\eqref{eq:wigner}--\eqref{eq:translation})
with the current observation coefficients:
{\setlength\abovedisplayskip{1pt}
\setlength\belowdisplayskip{1pt}
\begin{equation}
c_{\ell m}^{(t)}
=
\tilde{c}_{\ell m}^{(t)}
+
b_{\ell m}^{(t)}.
\label{eq:update}
\end{equation}}
This is the log-space Bayesian update under conditionally independent
observations: projecting the log-posterior sum onto the SH basis converts
log-likelihood addition in function space into coefficient addition — an
exact operation given the linearity of the SH projection.
Each \(b_{\ell m}^{(t)}\) contributes the current frame's directional
log-likelihood, while \(\tilde{c}_{\ell m}^{(t)}\) summarises all past
information propagated into the current viewer-centric frame.
The posterior \(\mathbf{c}^{(t)}\) is then stored and serves as the
input to the next update cycle.
\vspace{-3pt}

\noindent\textbf{Direction decoding.}
After updating the belief, we recover a single direction estimate via
the \emph{linear directional statistic} of the log-belief field — the
first moment of \(f_t\) viewed as a signed weight function on
\(\mathbb{S}^2\):
{\setlength\abovedisplayskip{1pt}
\setlength\belowdisplayskip{1pt}
\begin{equation}
\hat{\boldsymbol{\omega}}^{*}_{t}
=
\frac{\mathbf{m}_t}{\|\mathbf{m}_t\|},
\qquad
\mathbf{m}_t
=
\int_{\mathbb{S}^2} f_t(\boldsymbol{\omega})\,\boldsymbol{\omega}\,d\Omega.
\label{eq:decode}
\end{equation}}
Two properties make this a principled estimator in log-space.
\emph{First}, \(\mathbf{m}_t\) is invariant to any additive constant in
\(f_t\): since \(\int_{\mathbb{S}^2}\boldsymbol{\omega}\,d\Omega=\mathbf{0}\)
by symmetry, adding a constant \(K\) to \(f_t\) shifts \(\mathbf{m}_t\)
by \(K\cdot\mathbf{0}=\mathbf{0}\), leaving \(\hat{\boldsymbol{\omega}}^{*}_{t}\)
unchanged. This eliminates dependence on the arbitrary normalisation
constant inherent in any log-posterior, resolving the key concern about
computing a directional statistic from a log-density directly.
\emph{Second}, under a unimodal, strongly concentrated log-belief,
\(\hat{\boldsymbol{\omega}}^{*}_{t}\) coincides with the MAP direction:
because \(\exp\) is strictly monotone,
\(\arg\max_{\boldsymbol{\omega}} f_t(\boldsymbol{\omega})
=\arg\max_{\boldsymbol{\omega}}\exp(f_t(\boldsymbol{\omega}))\),
so the mode of \(f_t\) and the mode of the corresponding unnormalised
posterior density are identical, and the linear statistic \(\mathbf{m}_t\),
concentrated near this peak, recovers the MAP estimate without requiring
explicit exponentiation or normalisation.

\begin{wrapfigure}{r}{0.72\textwidth}
    \vspace{-12pt}
    \centering
    \includegraphics[width=0.70\textwidth]{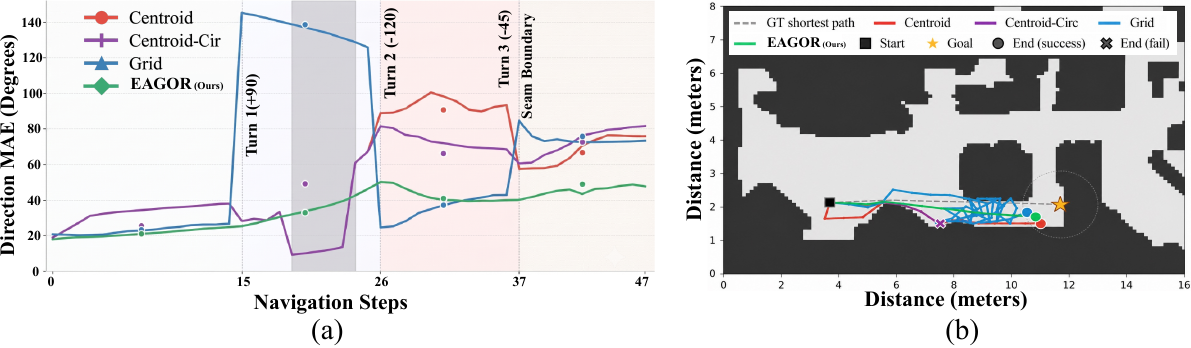}
    \vspace{-5pt}
    \caption{\textbf{Embodied omni-directional reasoning in Habitat-Sim:} 
(a) \textit{Waypoint Following}
(b) \textit{Map-Free Navigation}.}
    \label{fig:nav_analysis}
    \vspace{-10pt}
\end{wrapfigure}

Since the degree-1 SH coefficients encode the mean direction analytically,
\(\mathbf{m}_t\) is computed at \(O(1)\) cost directly in coefficient space,
without grid search or gradient ascent (\textcolor{Bittersweet}{\textit{Suppl.~Mat.~Sec.~C.1}}).
The resultant length \(R_t=\|\mathbf{m}_t\|\) serves as an unnormalised
concentration measure: by the same shift-invariance argument, \(R_t\) is
unaffected by additive constants in \(f_t\) and provides a reliable
relative confidence signal across timesteps — a larger \(R_t\) corresponds
to a more sharply peaked log-belief and a more confident direction estimate
(\textcolor{Bittersweet}{\textit{Suppl.~Mat.~Sec.~C.2}}).

Together, Eqs.~\eqref{eq:sh_belief}--\eqref{eq:decode} define a recursive
Bayesian filter on \(\mathbb{S}^2\): the current panorama produces a
directional log-likelihood observation; the SH coefficients store the
running log-posterior; Wigner-D rotation and, during occlusion, FOE/FOC
advection propagate the previous posterior into the current viewer-centric
frame; and the linear directional statistic of the log-belief returns
the final egocentric target direction.

%% file: sections/experiments.tex
\vspace{-10pt}
\section{Experiments and Evaluation}
\vspace{-10pt}

\begin{figure*}[t!]
    \centering
    \vspace{-10pt}
    \includegraphics[width=0.95\linewidth]{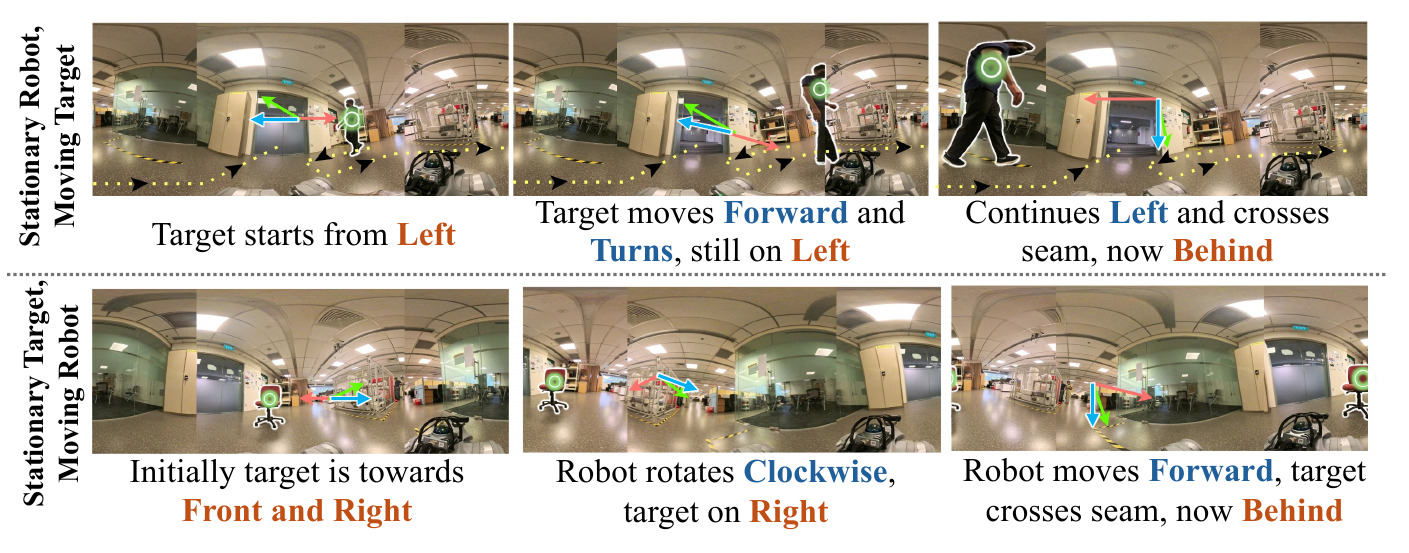}
    \vspace{-10pt}
    \caption{\textbf{Direction Estimation under Stationary and Mobile Operation.} Top: stationary robot with moving target, ($\cdots$) is the target trajectory. Bottom: moving robot with stationary target. In each frame, (\textcolor{ForestGreen}{$\rightarrow$}) denotes the spherical directional prediction, (\textcolor{red}{$\rightarrow$}) the ERP-based VLM baseline, and (\textcolor{ForestGreen}{$\square$}) the accumulated directional belief field. $\bigcirc=$ Target, and (\textcolor{cyan}{$\rightarrow$}) denotes the Ground Truth.}
    \label{fig:qualitative_real}
    \vspace{-15pt}
\end{figure*} 

\subsection{Experimental Settings and Implementation Details}

\begin{wraptable}{r}{0.65\textwidth}
\centering
\small
\caption{\textbf{Results for \textit{Active Visual Search}} on HOS and 
OSR-Bench. Best accuracy in \textbf{bold} and second-best \underline{underlined}.}
\label{tab:hos_results}
\vspace{-5pt}
\resizebox{0.63\textwidth}{!}{%
\begin{tabular}{l c c c c c c}
\toprule
& & \multicolumn{4}{c}{\textbf{Humanoid Object Search}} 
& \\
\cmidrule(lr){3-6}
\textbf{Method} & \textbf{Params} 
& \textbf{Overall} & \textbf{Easy} 
& \textbf{Medium} & \textbf{Hard}
& \textbf{OSR-Bench} \\
\midrule
LLaVA-v1.5 \cite{liu2023visualinstructiontuning}             & 3B   & 13.60 & 39.20 & 10.60 & 11.20 & 16.9 \\
Janus-Pro \cite{chen2025janusprounifiedmultimodalunderstanding}               & 3B   &    -- &    -- &    -- &    -- & 15.9 \\
Qwen2.5-VL \cite{bai2023qwentechnicalreport}               & 72B  &    -- &    -- &    -- &    -- & 18.1 \\
\arrayrulecolor{orange!60}\midrule
Gemma-3 \cite{gemmateam2025gemma3technicalreport}        & 4B   & 20.60 & 29.50 & 19.20 & 17.50 & 12.1 \\
\rowcolor{gray!10}
\quad + EAGOR            & 4B   & 24.80 & 40.20 & 24.45 & 18.90 & 17.2 \\
\arrayrulecolor{orange!60}\midrule
Gemma-3        & 12B  & 25.67 & 35.10 & 24.00 & 22.40 & 15.6 \\
\rowcolor{gray!10}
\quad + EAGOR            & 12B  & \underline{36.55} & \textbf{56.70} 
                                & \underline{38.30} & \underline{31.65} & 20.2 \\
\arrayrulecolor{orange!60}\midrule
Qwen2.5-VL     & 3B   & 20.71 & 32.00 & 22.10 & 16.40 & 12.7 \\
\rowcolor{gray!10}
\quad + EAGOR            & 3B   & 26.40 & 45.10 & 21.70 & 20.80 & \underline{20.4} \\
\arrayrulecolor{orange!60}\midrule
Qwen2.5-VL    & 7B   & 27.24 & 37.60 & 27.80 & 23.10 & 16.8 \\
\rowcolor{gray!10}
\quad + EAGOR            & 7B   & \textbf{40.10} & \underline{54.20} 
                                & \textbf{39.40} & \textbf{34.60} & \textbf{25.2} \\
\arrayrulecolor{black}\bottomrule
\end{tabular}%
}
\end{wraptable}

\vspace{-7pt}
We evaluate EAGOR as a training-free, model-agnostic framework for 
embodied omni-directional reasoning. Our evaluation is designed to 
validate two core claims: that SH-BF maintains geometrically consistent 
directional belief under egocentric motion, and that this geometric 
correctness translates to improved task performance across spatial tasks like navigation 
and active visual search without task-specific training.
\textit{Embodied Omni-directional Reasoning in Simulation} validates the geometric modeling of our framework in Habitat~\cite{savva2019habitatplatformembodiedai,
ramakrishnan2021habitatmatterport3ddatasethm3d}: \textbf{(a)~Waypoint 
Following} tests directional belief consistency under egocentric motion 
and seam crossings, while \textbf{(b)~Map-Free Navigation} evaluates 
whether geometric correctness translates to reliable closed-loop 
navigation without a map.
\textbf{(c)~Active Visual Search} evaluates EAGOR on \textit{object search under partial panoramic observation} in H*-Bench HOS~\cite{yu2025thinking360deghumanoidvisual} and \textit{multi-object egocentric spatial reasoning} in OSR-Bench~\cite{dongfang2025multimodallargelanguagemodels}. 
Although these benchmarks were originally designed for supervised direction regression, EAGOR adapts to them without any task-specific training by interpreting each partial observation as a directional likelihood on the sphere.
For \textit{Real-World Validation}, we deploy EAGOR on a Unitree-Go2 
equipped with an Insta360-X3 camera for \textbf{(d)~Target Direction 
Estimation} under stationary and mobile operation, verifying that 
geometric correctness transfers to physical deployment under real 
sensor noise.
At each timestep, the agent moves a fixed step along $\hat{\omega}^*_t$; 
the resulting parallax remains within single-frame observation noise, 
avoiding explicit translational compensation.
\vspace{-3pt}

\noindent\textbf{Baselines and Implementation Details.}
For HOS, OSR-Bench, and real-world evaluation, we compare \textit{Qwen2.5+EAGOR} and \textit{Gemma-3+EAGOR} against fine-tuned and zero-shot VLM baselines. 
For navigation-based experiments, we compare EAGOR against three ERP-space baselines: \textit{Centroid}, \textit{Centroid-Circ}, and \textit{Grid}. 
All experiments, including the baselines, are conducted under same setting on a system with $4\times$ NVIDIA RTX 5090 GPUs, each with 32GB of VRAM.

\vspace{-3pt}
Additional experimental and baseline details are provided in \textcolor{Bittersweet}{\textit{Suppl.\ Mat.\ Sec.~2}}.


\vspace{-7pt}
\subsection{Results and Discussion}
\label{sec:result}
\vspace{-7pt}
We organize our evaluation around three experimental questions --
\vspace{-2pt}
\begin{center}
\vspace{-7pt}
\colorbox{gray!10}{
\begin{minipage}{0.96\linewidth}
\textbf{EQ-1:} Can EAGOR adapt and improve end-to-end active visual search by converting VLM observations into a continuous directional belief across different VLM backbones ? \\[2pt]
\textbf{EQ-2:} Does EAGOR provide geometrically equivariant directional reasoning under egocentric motion, seam crossings, and map-free navigation in simulated 360$^\circ$ environments? \\[2pt]
\textbf{EQ-3:} Can EAGOR maintain consistent omni-directional reasoning in real-world settings ?
\end{minipage}
}
\vspace{-5pt}
\end{center}

\textbf{\underline{Active Visual Search}.}
Table~\ref{tab:hos_results} presents the quantitative results on H*-Bench Humanoid Object Search (HOS) and OSR-Bench.
For HOS, we evaluate EAGOR across frozen VLM backbones ranging from 3B to 12B parameters under a fully zero-shot setting.
Across all backbones, EAGOR consistently improves target-direction estimation, achieving roughly $1.7\times$--$2.2\times$ gains over the corresponding standalone models.
Notably, \textit{Qwen2.5-VL-7B} with \textit{EAGOR} surpasses the larger standalone \textit{Gemma-3-12B}, showing that explicit spherical belief reasoning can outweigh increased backbone capacity.
We further evaluate \textit{EAGOR} on 2,000+ OSR-Bench on multi-target relative-direction queries. As shown in Table~\ref{tab:hos_results}, \textit{EAGOR+Qwen2.5-VL-7B} achieves \textbf{25.2\%}, outperforming its canonical backbone (\textbf{16.8\%}), the $10\times$ larger \textit{Qwen2.5-VL-72B} (\textbf{18.1\%}). 
EAGOR is also efficient, converging in $1.4\times$ fewer views than the  baseline, with accuracy improving monotonically from $8.5\%$ to $40.1\%$ across turns. Refer to \textcolor{Bittersweet}{\textit{Suppl.\ Mat.\ Sec.~3}} for qualitative results and additional analyses.

\begin{center}
\vspace{-7pt}
\colorbox{gray!10}{
\begin{minipage}{0.96\linewidth}
\textcolor{DeepPink}{\textbf{[A-1]}}
By transforming VLM observations into a persistent directional belief, EAGOR consistently improves embodied omni-directional reasoning across frozen VLM backbones, enabling smaller models to outperform substantially larger ones without task-specific training.
\end{minipage}
}
\vspace{-5pt}
\end{center}

\underline{\textbf{Waypoint Following.}}
Table~\ref{tab:waypoint} presents the mean angular error across (MAE) four sequential trajectory segments (\textit{L1--L4}), with increasingly challenging turns between segments, together with robustness under occlusion, seam crossings, and temporal inconsistency.
The raw VLM centroid degrades substantially under rotation and seam crossings, while Cent-Circ only partially alleviates wrap-around errors.
Grid introduces temporal accumulation, but remains unstable under rotation and occlusion.
In contrast, EAGOR achieves the lowest error across all segments and robustness conditions, while reducing temporal inconsistency from $7.3^\circ\text{--}19.2^\circ/\mathrm{step}$ to $2.6^\circ/\mathrm{step}$.
Fig.~\ref{fig:nav_analysis}\textcolor{DeepPink}{a} shows that baseline errors spike after turns and near seam boundaries, while EAGOR maintains overall low angular error ($-34.4^\circ$) while handling seam crossings resulting in more accurate path following.
\vspace{-3pt}

\begin{wrapfigure}{r}{0.4\linewidth}
\centering
    \vspace{-10pt}
    \centering
    \includegraphics[width=.99\linewidth]{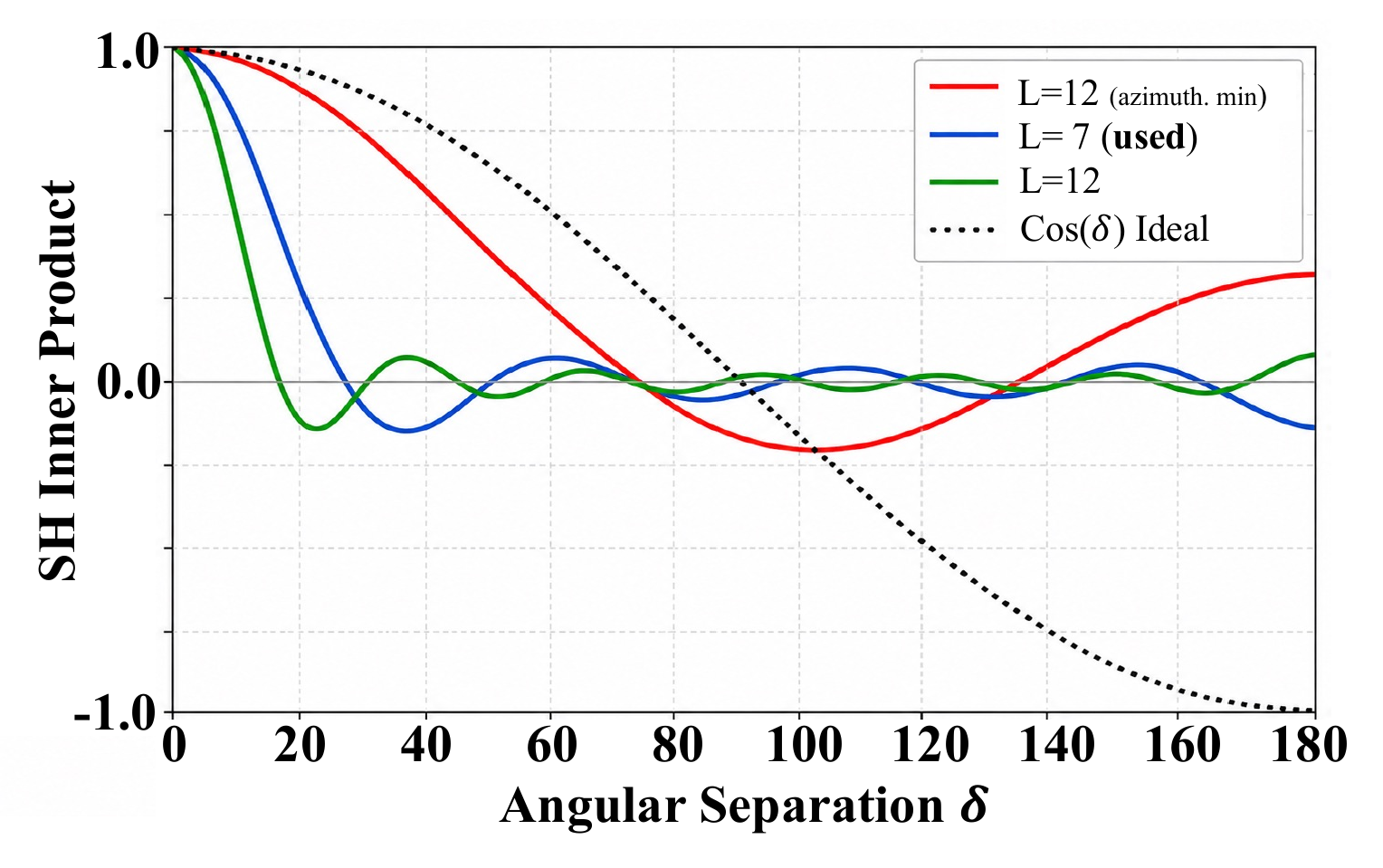}
    \vspace{-18pt}
    \captionof{figure}{\textbf{Effect of SH bandlimit.}}
    \label{fig:ablation}
\vspace{-15pt}
\end{wrapfigure}

\underline{\textbf{Map-Free Navigation.}}
Table~\ref{tab:nav} evaluates closed-loop navigation with Qwen2.5-VL-7B from near-opposite initial headings, where the agent executes the predicted direction at each step.
EAGOR achieves the best performance across all metrics, with the highest success rate and path efficiency ($94\%$ SR and $56.8$ SPL), the fewest steps ($50.2$), and the lowest angular error ($33.8^\circ$).
Although Cent-Circ improves seam handling over Centroid, its lower success rate indicates that per-frame wrap-around correction is insufficient for reliable control.
Similarly, Grid matches Centroid in success rate but produces less efficient trajectories, showing that temporal accumulation alone does not ensure geometrically consistent actions.
EAGOR further achieves substantially stronger seam robustness ($70.6$). 
As illustrated in Fig.~\ref{fig:qualitative} and Fig.~\ref{fig:nav_analysis}\textcolor{DeepPink}{b}, EAGOR
enables efficient and successful closed-loop navigation.
Additional results are provided in the \textcolor{Bittersweet}{\textit{Suppl. Mat. Sec.~3}}. 

\begin{center}
\vspace{-5pt}
\colorbox{gray!10}{
\begin{minipage}{0.96\linewidth}
\textcolor{DeepPink}{\textbf{[A-2]}}
EAGOR, therefore, provides geometrically consistent directional reasoning in simulated $360^\circ$ environments by maintaining a persistent spherical belief that remains stable under egocentric motion and seam crossings, and translates into reliable map-free navigation.
\end{minipage}
}
\vspace{-5pt}
\end{center}

\underline{\textbf{Target Direction Estimation under Stationary and Mobile Operation.}}
Fig.~\ref{fig:qualitative_real} qualitatively evaluates directional estimation when the robot is stationary and while it moves to follow a target.
Across both settings, EAGOR maintains a coherent target direction despite real sensor noise, viewpoint changes, ego-motion, and temporary target disappearance. Additional results in  \textcolor{Bittersweet}{\textit{Suppl. Mat. Sec.~3.}}

\begin{center}
\vspace{-8pt}
\colorbox{gray!10}{
\begin{minipage}{0.96\linewidth}
\textcolor{DeepPink}{\textbf{[A-3]}}
EAGOR maintains consistent omni-directional reasoning in real-world indoor environments, preserving a coherent target-direction belief under both stationary and mobile robotic operation despite sensor noise, viewpoint changes, and ego-motion.
\end{minipage}
}
\vspace{-8pt}
\end{center}


\vspace{-5pt}
\subsection{Ablation on SH Bandlimit and Runtime Analysis}
\vspace{-8pt}


The SH bandlimit $L$ balances angular precision and sidelobe ringing. As shown in Fig.~\ref{fig:ablation}, larger $L$ yields sharper localization but stronger oscillations at wider angular separations. We use $L=7$, which provides ${\approx}25^\circ$ angular resolution while avoiding excessive Gibbs ringing. 
\vspace{-3pt}

\textbf{Running Time.} EAGOR requires $0.6\times$ more inference time than the baseline ($\sim1.6$ vs.\ $\sim1.0$s with Qwen2.5-VL-7B), with almost \textit{no extra} GPU memory. This is marginal for a training-free approach and can be further reduced via optimizations~\cite{wu2024controlmllm, debnath2026llmindbioinspiredtrainingfreeadaptive}.

%% file: sections/conclusion.tex
\vspace{-7pt}
\section{Conclusion, Limitations, and Future Work}
\label{sec:conclusion}
\vspace{-10pt}

\paragraph{Conclusion.}
We introduced \textbf{EAGOR}, a training-free framework for embodied omni-directional reasoning that formulates dynamic target localization as recursive belief estimation on the sphere. By maintaining a continuous egocentric directional belief and propagating it consistently under motion, EAGOR enables geometry-aware reasoning beyond frame-wise panoramic localization. Experiments in simulation and on a real-world legged robot demonstrate the importance of spherical belief representations for robust embodied omni-directional perception.
\vspace{-10pt}

\begin{wrapfigure}{r}{0.4\linewidth}
    \vspace{-12pt}
    \centering
    \captionof{table}{\textbf{Failure mode analysis on HOS dataset} (Qwen2.5-VL-7B).}
    \vspace{-5pt}
    \label{tab:failure_analysis}
    \resizebox{\linewidth}{!}{%
    \begin{tabular}{lcc}
        \toprule
        \textbf{Failure Mode} & \textbf{Episodes} & \textbf{Success Rate (\%)} \\
        \midrule
        OCR / fine-grained text & 18.5 & 13.2 \\
        Rare target              & 16.9 & 3.6  \\
        Multi-instance confusion & 12.5 & 33.3 \\
        VLM false detection      & 11.9 & 13.6 \\
        \midrule
        \textit{Overall}         & \textit{59.9} & \textit{40.1} \\
        \bottomrule
    \end{tabular}%
    }
\vspace{-10pt}
\end{wrapfigure}
\paragraph{Limitations and Future Work.}
EAGOR uses VLM observations as directional likelihoods and is therefore sensitive to semantic ambiguity and perception errors. As shown in Tab.~\ref{tab:failure_analysis}, its main failure modes arise from rare targets, multi-instance confusion, and false detections. Nevertheless, its performance on multi-instance cases indicates that cross-view belief accumulation partially mitigates ambiguous observations. Future work will incorporate grounded detection for multi-instance disambiguation and translation-aware, multi-sensor formulations for parallax-aware reasoning. 
We will explore bio-inspired perception for improved efficiency~\cite{debnath2026llmind, liu2026vl2spike}.
Project will be made public.

\begin{table*}[t]
\vspace{-5pt}
\centering
\footnotesize

\begin{minipage}[t]{0.56\textwidth}
\centering
\caption{\textbf{Waypoint Following.} Mean angular error (MAE, $^\circ$, $\downarrow$) over four trajectory segments (\textit{L1--L4}). $\Delta$ denotes trajectory degradation; \textit{Occ} and \textit{Seam} report MAE under occlusion and seam crossing error ($^\circ$, $\downarrow$); \textit{TC} measures temporal inconsistency ($^\circ/\mathrm{step}$, $\downarrow$).}
\label{tab:waypoint}
\vspace{-3pt}

\resizebox{\textwidth}{!}{
\begin{tabular}{lcccccccc}
\toprule
Method & L1 & L2 & L3 & L4 & $\Delta$ & Occ & Seam & TC \\
\midrule
Centroid   & 25.5 & 49.4 & 90.9 & 66.7 & +40.3 & 31.1 & 59.1 & 11.6 \\
Cent-Circ  & 25.5 & 49.3 & 66.3 & 72.8 & +46.4 & 31.1 & 51.0 & 7.3 \\
Grid       & 23.2 & 138.9 & 37.3 & 75.9 & +52.3 & 135.2 & 42.4 & 19.2 \\
\textbf{EAGOR} & \textbf{21.1} & \textbf{31.1} & \textbf{40.9} & \textbf{47.1} 
& \textbf{+25.4} & \textbf{15.5} & \textbf{22.9} & \textbf{2.6} \\
\bottomrule
\end{tabular}
}
\end{minipage}
\hfill
\begin{minipage}[t]{0.41\textwidth}
\centering
\caption{\textbf{Map-Free Navigation.} Closed-loop navigation from near-opposite headings ($|\theta| \approx 180^\circ$). Metrics include success (SR), path efficiency (SPL), steps, MAE, and seam-crossing success.}
\label{tab:nav}
\vspace{-3pt}

\resizebox{\textwidth}{!}{
\begin{tabular}{lccccc}
\toprule
Method & SR$\uparrow$ & SPL$\uparrow$ & Steps$\downarrow$ & MAE$\downarrow$ & Seam$\uparrow$ \\
\midrule
Centroid   & 82 & 54.4 & 61.0 & 45.9 & 19.6 \\
Cent-Circ  & 62 & 41.5 & 77.8 & 46.5 & 33.5 \\
Grid       & 82 & 41.4 & 63.6 & 54.2 & 15.8 \\
\textbf{EAGOR} & \textbf{94} & \textbf{56.8} 
& \textbf{50.2} & \textbf{33.8} & \textbf{70.6} \\
\bottomrule
\end{tabular}
}
\end{minipage}
\vspace{-14pt}
\end{table*}
